\pdfoutput=1

\documentclass[11pt]{article}

\usepackage[preprint]{acl}
\usepackage[skins,most]{tcolorbox} 
\tcbuselibrary{breakable} 
\newtcolorbox[auto counter, number within=section, list type=subsubsection, list inside=toc]{sectionbox}[2][]{
colback=white!98!gray, colframe=black, 
colbacktitle=white!90!gray, coltitle=black, 
fonttitle=\bfseries,
title={#2}, 
list entry={Comment \thetcbcounter\quad}
}

\usepackage{booktabs}
\usepackage{subcaption}
\usepackage{times}
\usepackage{latexsym}
\usepackage{amsmath}
\usepackage{amssymb}
\usepackage{algorithm}
\usepackage{algpseudocode}
\usepackage{xcolor}
\usepackage{color}
\usepackage{listings}
\definecolor{customTeal}{RGB}{0, 128, 128}
\definecolor{emphasisColor}{RGB}{255, 0, 0} 
\usepackage{tabularx}
\usepackage{ragged2e}
\newcolumntype{Y}{>{\RaggedRight\arraybackslash}X} 
\usepackage[skins,most]{tcolorbox}
\usepackage{colortbl}
\usepackage{multirow} 
\usepackage{booktabs}
\usepackage{wrapfig}
\usepackage{tikz}
\usetikzlibrary{
    arrows.meta,
    shapes.geometric, 
    positioning
}
\usepackage{capt-of}
\usepackage[edges]{forest}  
\tikzset{
  my-box/.style={
    rectangle,
    draw=hidden-black,
    rounded corners,
    text opacity=1,
    minimum height=1.5em,
    minimum width=5em,
    inner sep=2pt,
    align=center,
    fill opacity=.5,
  },
  leaf/.style={
    my-box,
    minimum height=1.5em,
    fill=yellow!32,
    text=black,
    align=left,
    font=\normalsize,
    inner xsep=5pt,
    inner ysep=4pt,
    text width=45em, 
  },
  leaf2/.style={
    my-box,
    minimum height=1.5em,
    fill=purple!27,
    text=black,
    align=left,
    font=\normalsize,
    inner xsep=5pt,
    inner ysep=4pt,
  },
  leaf3/.style={
    my-box,
    minimum height=1.5em,
    fill=hidden-blue!57,
    text=black,
    align=left,
    font=\normalsize,
    inner xsep=5pt,
    inner ysep=4pt,
  },
  leaf4/.style={
    my-box,
    minimum height=1.5em,
    fill=green!14,
    text=black,
    align=left,
    font=\normalsize,
    inner xsep=5pt,
    inner ysep=4pt,
    },
    leaf5/.style={
        my-box,
        minimum height=1.5em,
        fill=orange!16,
        text=black,
        align=left,
        font=\normalsize,
        inner xsep=5pt,
        inner ysep=4pt,
    },
}

\usepackage{pifont}  
\definecolor{customgreen}{HTML}{16C47F}  
\definecolor{customred}{HTML}{C62300}   
\definecolor{hidden-red}{RGB}{205, 44, 36}
\definecolor{hidden-blue}{RGB}{194,232,247}
\definecolor{hidden-orange}{RGB}{243,202,120}
\definecolor{hidden-green}{RGB}{34,139,34}
\definecolor{hidden-pink}{RGB}{255,245,247}
\definecolor{hidden-black}{RGB}{20,68,106}
\definecolor{purple}{RGB}{144,153,196}
\definecolor{yellow}{RGB}{255,228,123}
\definecolor{hidden-yellow}{RGB}{255,248,203}
\definecolor{tkcolor}{RGB}{224,223,255}
\definecolor{hidden-draw}{RGB}{128,128,128}
\definecolor{darkblue}{rgb}{0, 0.40, 0.75} %
\definecolor{colorEssential}{HTML}{0D47A1} 
\definecolor{colorCommon}{HTML}{1E88E5}    
\definecolor{colorNiche}{HTML}{BDBDBD}      

\usepackage{float}

\usepackage{caption}

\usepackage[T1]{fontenc}

\usepackage[utf8]{inputenc}

\usepackage{microtype}

\usepackage{inconsolata}

\usepackage{enumitem} 
\usepackage{lipsum}
\usepackage{graphicx}

\NewDocumentCommand{\yi}
{ mO{} }{\textcolor{blue}{\textsuperscript{\textit{May}}\textsf{\textbf{\small[#1]}}}}
\NewDocumentCommand{\yuchen}
{ mO{} }{\textcolor{red}{\textsuperscript{\textit{Yuchen}}\textsf{\textbf{\small[#1]}}}}
\NewDocumentCommand\emojienvisage{}{\includegraphics[width=0.85em, height=0.85em]{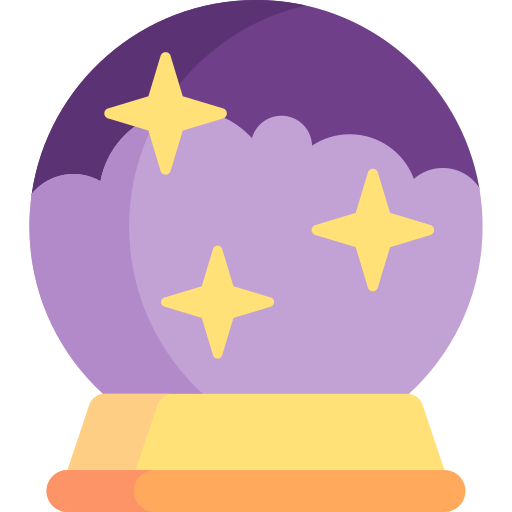}}

\title{\includegraphics[width=1em, height=1em]{pictures/envisage.png} Environment Scaling for Interactive  Agentic\\ Experience Collection: A Survey}

\DeclareSymbolFont{extraup}{U}{zavm}{m}{n}
\DeclareMathSymbol{\vardiamond}{\mathalpha}{extraup}{87}
\author{Yuchen Huang$^{\heartsuit}$ Sijia Li$^{\heartsuit}$ Minghao Liu$^{\heartsuit}$ Wei Liu$^{\spadesuit}$ Shijue Huang$^{\heartsuit}$ Zhiyuan Fan$^{\heartsuit}$\\  \textbf{Hou Pong Chan$^{\clubsuit}$ Yi R. (May) Fung$^{\heartsuit}$}\\
$^{\heartsuit}$Hong Kong University of Science and Technology \\ 
$^{\spadesuit}$King's College London ~$^{\clubsuit}$DAMO Academy, Alibaba Group\\ 
\{yhuanggn, yrfung\}@cse.ust.hk \\
}

\begin{document}
\maketitle

\begin{abstract}
LLM-based agents can autonomously accomplish complex tasks across various domains. However, to further cultivate capabilities such as adaptive behavior and long-term decision-making, training on static datasets built from human-level knowledge is insufficient. These datasets are costly to construct and lack both dynamism and realism. A growing consensus is that agents should instead interact directly with environments and learn from experience through reinforcement learning. We formalize this iterative process as the \textit{Generation-Execution-Feedback (GEF) loop}, where environments generate tasks to challenge agents, return observations in response to agents' actions during task execution, and provide evaluative feedback on rollouts for subsequent learning. Under this paradigm, environments function as indispensable producers of experiential data, highlighting the need to scale them toward greater complexity, realism, and interactivity. In this survey, we systematically review representative methods for environment scaling from a pioneering environment-centric perspective and organize them along the stages of the \textit{GEF loop}, namely task generation, task execution, and feedback. We further analyze implementation frameworks, challenges, and applications, consolidating fragmented advances and outlining future research directions for agent intelligence.
\footnote{We provide a GitHub repository with real-time updates on this topic: \url{https://github.com/lukahhcm/Awesome_Scaling_Environments}.}
\end{abstract}

\section{Introduction}
\label{sec:intro}
The rapid progress of large language models (LLMs) has catalyzed a transformative shift in artificial intelligence, precipitating a surge of research on LLM-based agents~\citep{luo2025largelanguagemodelagent, xi2025rise}. 
Such agents inherit strong reasoning and task-decomposition capabilities from their base models and, when augmented with modules for tool use and memory, can execute actions, interact with real or simulated environments, accumulate experience over time, and progressively improve their own behavior. This design has achieved remarkable progress across diverse domains, including automated coding~\citep{qwen3technicalreport, anthropic2025claude4}, interactive web navigation~\citep{openai2025introducing,he2025advancing}, tool use~\citep{zhang2025agentorchestra,anthropic2024mcp}, and deep research~\citep{tongyidr, openai2025deepresearch, google2024geminideepresearch}.

\begin{figure}[t]
\centering
\includegraphics[width=\columnwidth]{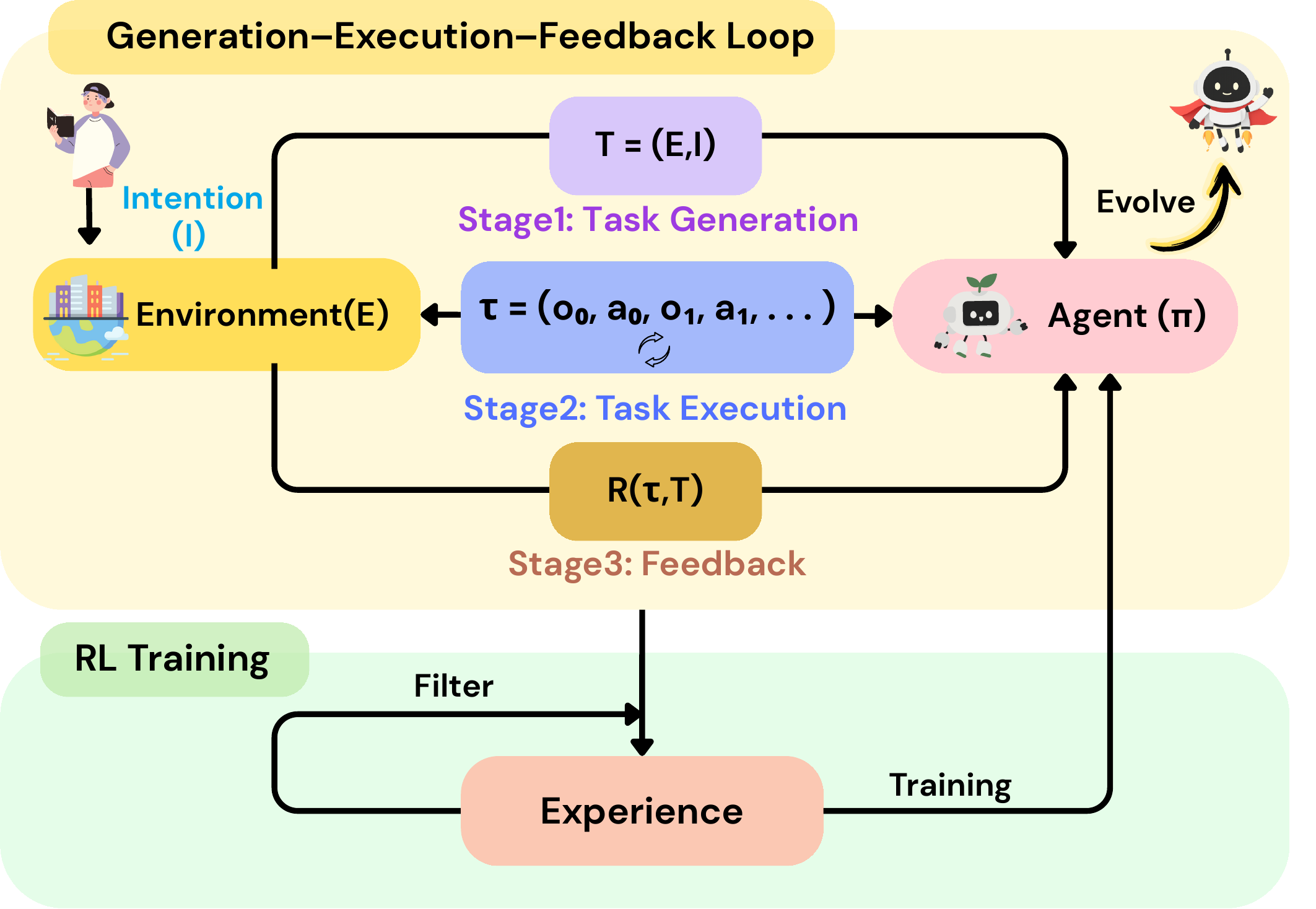}
\vspace{-0.6em}
\caption{Experience arises from the \textit{Generation-Execution-Feedback (GEF) loop}, where environments generate tasks, agents execute them, and environments evaluate and filter useful experience for RL training.}
\label{fig:loop}
\end{figure}

\tikzstyle{my-box}=[
    rectangle,
    draw=hidden-draw,
    rounded corners,
    text opacity=1,
    minimum height=1.5em,
    minimum width=5em,
    inner sep=2pt,
    align=center,
    fill opacity=.5,
]

\tikzstyle{level1-style}=[
    my-box,
    fill=gray!5,
    text=black,
    align=left,
    font=\scriptsize,
    inner xsep=5pt,
    inner ysep=3pt,
]

\tikzstyle{level2-style}=[
    my-box,
    fill=gray!10,
    text=black,
    align=left,
    font=\scriptsize,
    inner xsep=5pt,
    inner ysep=3pt,
]

\tikzstyle{level3-style}=[
    my-box,
    fill=gray!15,
    text=black,
    align=left,
    font=\scriptsize,
    inner xsep=5pt,
    inner ysep=3pt,
]

\tikzstyle{leaf}=[
    my-box, 
    minimum height=1.5em,
    fill=yellow!32.0, 
    text=black, 
    align=left,
    font=\scriptsize,
    inner xsep=8pt,
    inner ysep=4pt,
]
\tikzstyle{leaf2}=[
    my-box, 
    minimum height=1.5em,
    fill=purple!27, 
    text=black, 
    align=left,
    font=\scriptsize,
    inner xsep=8pt,
    inner ysep=4pt,
]
\tikzstyle{leaf3}=[
    my-box, 
    minimum height=1.5em,
    fill=hidden-blue!57, 
    text=black, 
    align=left,
    font=\scriptsize,
    inner xsep=8pt,
    inner ysep=4pt,
]
\tikzstyle{leaf4}=[
    my-box, 
    minimum height=1.5em,
    fill=green!14, 
    text=black, 
    align=left,
    font=\scriptsize,
    inner xsep=8pt,
    inner ysep=4pt,
]
\tikzstyle{leaf5}=[
    my-box, 
    minimum height=1.5em,
    fill=orange!16, 
    text=black, 
    align=left,
    font=\scriptsize,
    inner xsep=8pt,
    inner ysep=4pt,
]

\begin{figure*}[t]
    \centering
    \resizebox{\textwidth}{!}
    {
        \begin{forest}
            forked edges,
            for tree={
                grow=east,
                reversed=true,
                anchor=base west,
                parent anchor=east,
                child anchor=west,
                base=left,
                font=\small,
                rectangle,
                draw=hidden-draw,
                rounded corners,
                align=left,
                minimum width=4em,
                edge+={darkgray, line width=1pt},
                s sep=3pt,
                inner xsep=5pt,
                inner ysep=3pt,
                ver/.style={rotate=90, child anchor=north, parent anchor=south, anchor=center},
            },
            where level=1{text width=4.6em,font=\scriptsize,level1-style}{},
            where level=2{text width=5.5em,font=\scriptsize,level2-style}{},
            where level=3{text width=5.7em,font=\scriptsize,level3-style}{},
            where level=4{text width=6.1em,font=\scriptsize,}{},
            [
                Scaling Environments for Agents, ver
                [
                    Stage 1: Task \\ Generation~(\S\ref{sec:task_gen})
                    [
                        Complexity~(\S\ref{ssec:task-complexity})
                        [
                            \textit{e.g.,}
                            ToolBench \citep{qin2023toolllm}{,} 
                            BFCL V3 \citep{patil2025bfcl}{,} 
                            $\tau$-bench \citep{yao2024tau}{,} 
                            $c^3$-bench \citep{yu2025c}{,} \\
                            ToolHop~\citep{ye2025toolhop}{,}                    
                            TaskCraft~\citep{shi2025taskcraft}{,}
                            WebDancer~\citep{wu2025webdancerautonomousinformationseeking}{,}
                            WebShaper~\citep{tao2025webshaperagenticallydatasynthesizing}{,}\\
                            WebSailor~\citep{li2025websailornavigatingsuperhumanreasoning}{,}
                            WebExplorer~\citep{liu2025webexplorer}{,}
                            WebLeaper~\citep{tao2025webleaper}{,}
                            \textit{etc.}
                            , leaf3, text width=32.0em
                        ]
                    ]
                    [
                        Dynamic ~(\S\ref{ssec:task-dynamic})
                        [
                            \textit{e.g.,}
                            EvoCurr~\citep{cheng2025evocurr}{,}
                            WebRL~\citep{qiwebrl}{,}
                            AgentGym~\citep{xi2024agentgym}{,}
                            AgentGen~\citep{hu2025agentgen}{,} \\
                            R-Zero~\citep{huang2025r}{,}
                            EnvGen~\citep{zala2024envgen}{,}
                            ARE~\citep{andrews2025arescalingagentenvironments}{,}
                            RLVE-Gym~\citep{zeng2025rlve}{,}
                            \textit{etc.}
                            , leaf3, text width=32.0em
                        ]
                    ]
                    [
                        Diversity ~(\S\ref{ssec:task-diversity})
                        [
                            \textit{e.g.,}
                            TaskCraft~\citep{shi2025taskcraft}{,}
                            ARE~\citep{andrews2025arescalingagentenvironments}{,}
                            AutoEnv~\citep{zhang2025autoenv}{,} 
                            AgentGen~\citep{hu2025agentgen}{,} \\
                            AgentGym \citep{xi2024agentgym}{,} 
                            AgentBank \citep{song2024agentbank}{,} 
                            SWE-Gym \citep{pan2025trainingsoftwareengineeringagents}{,}
                            R2E-Gym \citep{jain2025r2egymproceduralenvironmentshybrid}{,}\\
                            SWE-Smith \citep{yang2025swesmithscalingdatasoftware}{,}
                            RLVE-Gym~\citep{zeng2025rlve}{,}
                            \textit{etc.}
                            , leaf3, text width=32.0em
                        ]
                    ]
                ]
                [
                    Stage 2: Task \\Execution~(\S\ref{sec:task_exe})
                    [
                        Interactivity~(\S\ref{ssec:task-interactivity})
                        [
                            \textit{e.g.,}
                            RandomWorld~\citep{sullivan2025procedural}{,}  
                            AppWorld~\citep{trivedi2024appworldcontrollableworldapps}{,}
                            BrowseMaster \citep{pang2025browsemaster}{,}\\
                            MCP-Bench \citep{wang2025mcp}{,} 
                            OSWorld-MCP\citep{jia2025osworldmcp}{,}
                            WebShop~\citep{yao2023webshopscalablerealworldweb}{,}
                            FTRL \citep{ye2025feedbackdriven}{,}\\
                            AndroidWorld~\citep{rawles2025androidworld}{,}
                            MiroThinker \citep{miromindteam2025mirothinker}{,}
                            AgentScaler \citep{fang2025generalagentic}{,}
                            \textit{etc.}
                            , leaf, text width=32.0em
                        ]
                    ]
                    [
                        Realism~(\S\ref{ssec:task-realism})
                        [
                            \textit{e.g.,}
                            Tongyi DR \citep{tongyidr}{,}
                            AgentScaler \citep{fang2025generalagentic}{,}
                            FTRL \citep{ye2025feedbackdriven}{,}\\
                            WebShop \citep{yao2023webshopscalablerealworldweb}{,}
                            WebArena \citep{zhou2023webarena}{,}
                            OSWorld \citep{xie2024osworld}{,}
                            OSWorld-G \citep{xie2025scalingcomputerusegroundinguser}{,}\\
                            $\tau^2$-bench \citep{barres2025tau2bench}{,}
                            AgentSociety \citep{zhang2025parallelized}{,}
                            ARE~\citep{andrews2025arescalingagentenvironments}{,}
                            \textit{etc.}
                            , leaf, text width=32.0em
                        ]
                    ]   
                ]
                [
                    Stage 3:\\ Feedback ~(\S\ref{sec:feedback})
                    [
                        Density ~(\S\ref{ssec:feedback-density})
                        [
                            \textit{e.g.,}
                            VerlTool \citep{jiang2025verltool}{,}
                            Agent-RLVR \citep{da2025agent}{,}
                            Toolrl \citep{qian2025toolrl}{,}
                            SRM \citep{ma2025steplevelrewardmodelsrewarding}{,}\\
                            ThinkPRM \citep{khalifa2025processrewardmodelsthink}{,}
                            Web-Shepherd \citep{chae2025webshepherd}{,}
                            PR-Clip-Delta \citep{gao2024designing}{,}                         
                            \textit{etc.}
                            , leaf5, text width=32.0em
                        ]
                    ]
                    [
                        Granularity ~(\S\ref{ssec:feedback-granularity})
                        [
                            \textit{e.g.,}
                            AdaCtrl \citep{huang2025adactrl}{,}
                            Toolrl \citep{qian2025toolrl}{,}
                            R1-Searcher \citep{song2025r1}{,}
                            RaR \citep{gunjal2025rubricsrewardsreinforcementlearning}{,}\\
                            Rubicon \citep{huang2025reinforcement}{,}
                            RuscaRL \citep{zhou2025breaking}{,}
                            DR Tulu \citep{shao2025drtulu}{,}
                            \textit{etc.}
                            , leaf5, text width=32.0em
                        ]
                    ]
                    [
                        Automation~(\S\ref{ssec:feedback-automation})
                        [
                            \textit{e.g.,}
                            JudgeLRM \citep{chen2025judgelrmlargereasoningmodels}{,}
                            GenRM-CoT \citep{zhang2025generativeverifiersrewardmodeling}{,}
                            REWARDAGENT \citep{peng2025agenticrewardmodelingintegrating}{,} \\
                            ARMAP \citep{chen2025scalingautonomousagentsautomatic}{,} 
                            RaR \citep{gunjal2025rubricsrewardsreinforcementlearning}{,}
                            Rubicon \citep{huang2025reinforcement}{,}
                            \textit{etc.}
                            , leaf5, text width=32.0em
                        ]
                    ]
                    [
                        Objectivity~(\S\ref{ssec:feedback-objectivity})
                        [
                            \textit{e.g.,}
                            RM-7B \citep{su2025crossingrewardbridgeexpanding}{,}
                            General-Reasoner \citep{ma2025general}{,}
                            Rlpr \citep{yu2025rlpr}{,}
                            BRPO \citep{jia2025writingzerobridgegapnonverifiable}{,}\\
                            Nemotron-Crossthink \citep{akter2025nemotron}{,}
                            Omni-Thinker \citep{li2025omni}{,}
                            ARE~\citep{andrews2025arescalingagentenvironments}{,}
                            \textit{etc.}
                            , leaf5, text width=32.0em
                        ]
                    ]
                    [
                        Robustness~(\S\ref{ssec:feedback-robustness})
                        [
                            \textit{e.g.,}
                            ScoreDiff \citep{lin2024navigating}{,}
                            MONA \citep{farquhar2025monamyopicoptimizationnonmyopic}{,}
                            PAR \citep{fu2025reward}{,}
                            RRM \citep{liu2024rrm}{,}\\
                            InfoRM \citep{miao2024inform}{,}
                            RRM \citep{liu2024rrm}{,}
                            RM-7B \citep{su2025crossingrewardbridgeexpanding}{,}
                            Rubicon \citep{huang2025reinforcement}{,}\\
                            Trinity-RFT \citep{pan2025trinityrftgeneralpurposeunifiedframework}{,}
                            Tongyi DR
                            \citep{tongyidr}{,}
                            \textit{etc.}
                            , leaf5, text width=32.0em
                        ]
                    ]
                ]
            ]
        \end{forest}
    }
    \caption{GEF-aligned taxonomy of environment scaling with dimensions for \textit{Task Generation}, \textit{Task Execution}, and \textit{Feedback}. Representative works are illustrated as leaves on the branches.}
    \vspace{-10pt}
    \label{fig:scaling-environments-survey}
\end{figure*}
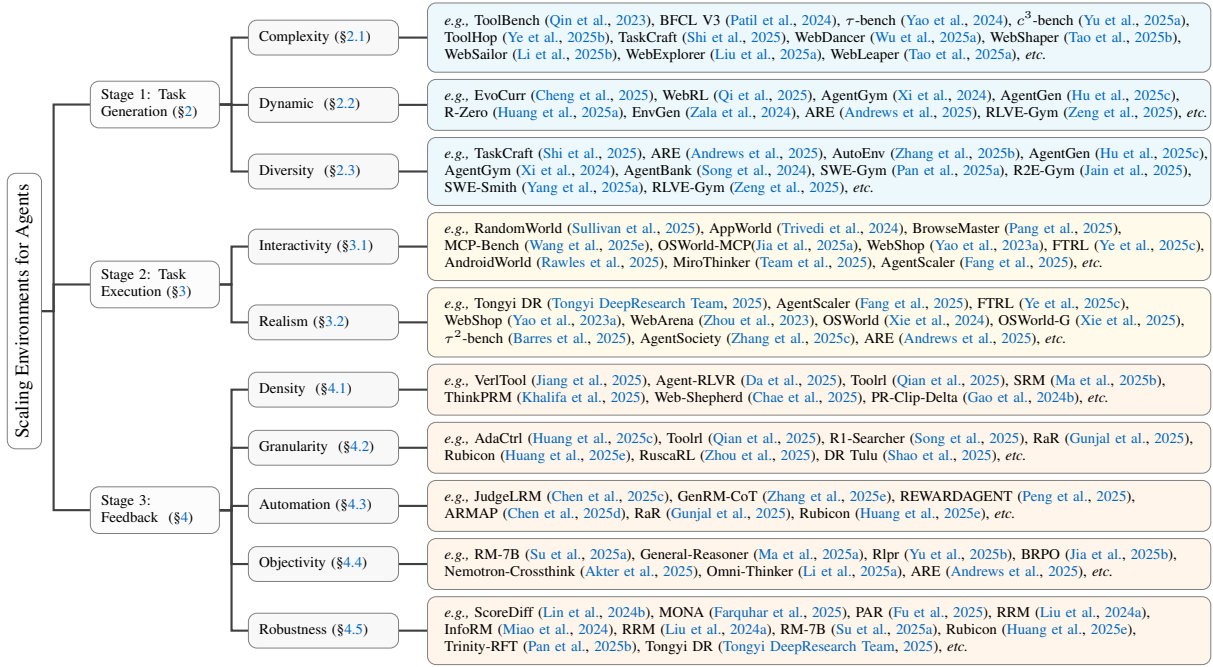

\begin{figure*}[t]
\includegraphics[width=\textwidth]{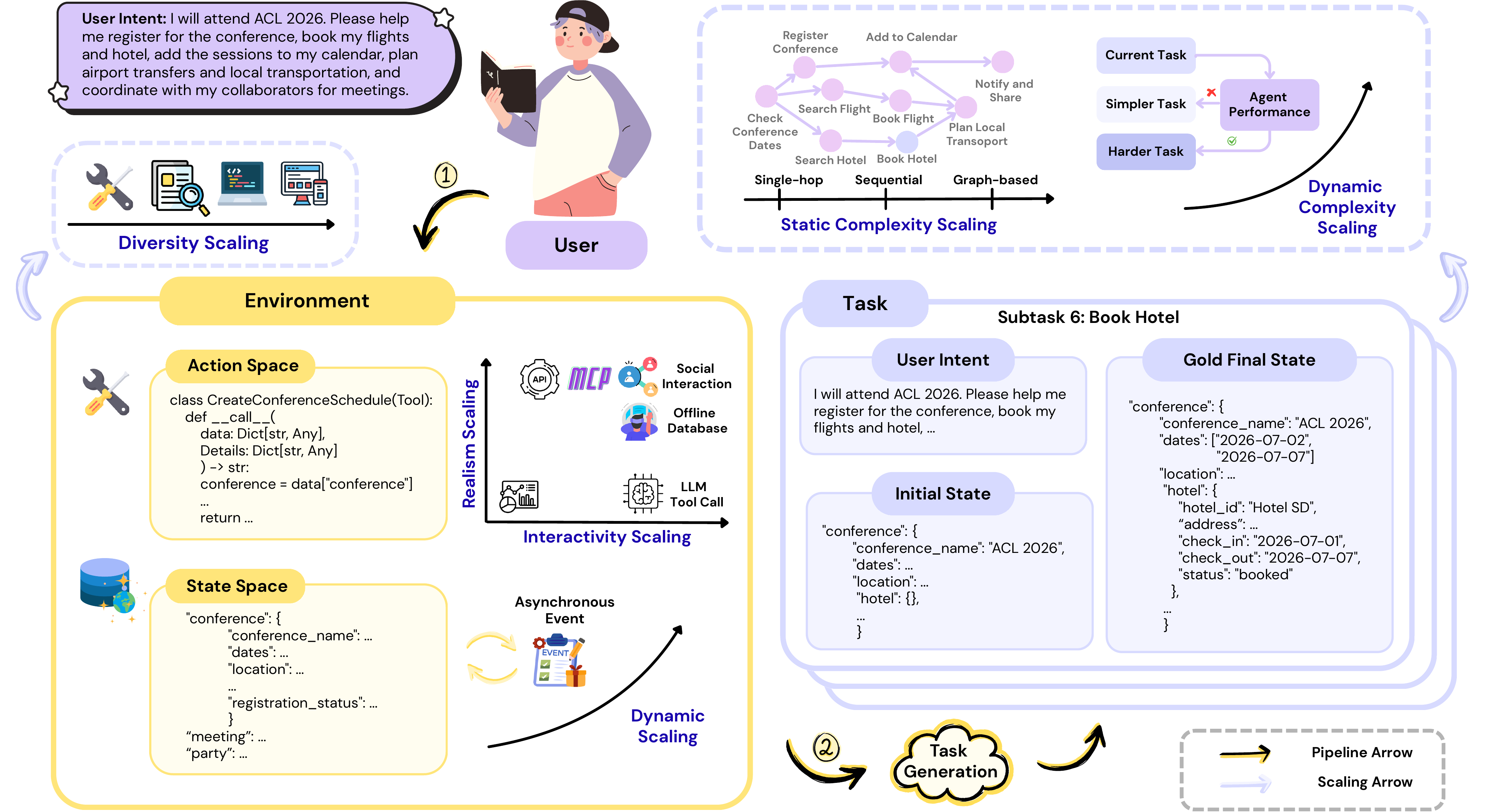}
\vspace{-0.6em}
\caption{Illustration of environment scaling in the \textbf{task generation} and \textbf{task execution} stages, using the example of conference scheduling. Given a user intent, the environment produces a set of tasks for the agent to complete. Scaling in the task generation stage covers \textit{complexity scaling}, \textit{dynamic scaling}, and \textit{diversity scaling}, while in the task execution stage scaling encompasses \textit{interactivity scaling} and \textit{realism scaling}.}
\label{fig:task}
\end{figure*}

However, as agent capabilities continue to evolve, it is infeasible to attain intelligence beyond the human-level merely by supervised fine-tuning (SFT) pretrained models on static datasets~\citep{huang2025r,su2025crossingrewardbridgeexpanding, zhao2025absolutezeroreinforcedselfplay}. Such datasets are typically manually annotated or curated under human oversight, which makes them costly and labor-intensive to produce at scale, intrinsically bounded by human-level knowledge, and lacking realism and adaptability. By contrast, reinforcement learning provides a more aligned training paradigm \citep{tao2024surveyselfevolutionlargelanguage,zhang2025landscapeagenticreinforcementlearning}, where agents can explore in the environment, accumulate experience, and finally acquire new knowledge or skills. We formalize this interactive process as the \textit{Generation-Execution-Feedback (GEF) loop}, illustrated in Figure~\ref{fig:loop}. In each iteration, the environment first generates diverse tasks, then the agent executes them within the environment, producing action-observation trajectories. The environment subsequently evaluates these rollouts and retains useful experience for subsequent training. Repeated iterations of this loop progressively refine the policy and expand the agent's capabilities. In this paradigm, the environment is no longer a mere container for agents' activities but an active producer of experiential data, underscoring the growing need for scaling environments to create a more complex, realistic, and richly interactive world~\citep{camel2025scaling}\footnote{Unlike prior work \citep{gao2025survey}, we adopt a broad view of the environment: everything external to the current agent, including the state space, the executable action space, the design of feedback for interaction and evaluation, and the activities of users and other agents, is considered part of it. Illustrative examples are shown in Figures \ref{fig:task} and \ref{fig:feedback}.}.

Recent research has embraced this trend of scaling the environment from different perspectives. Systems like AgentGen~\citep{hu2025agentgen} and GEM~\citep{liu2025gemgymagenticllms} devise heterogeneous environments to increase the diversity of the generated tasks. R-Zero~\citep{huang2025r} proposes a challenger-solver framework that autonomously generates increasingly difficult tasks. RandomWorld~\citep{sullivan2025procedural} scales up the interactivity by procedural generation of diverse tools for agents to access. ARE~\citep{andrews2025arescalingagentenvironments} develops an event-driven environment that supports asynchronous interactions that conform to realistic settings. However, a systematic analysis that connects these research directions remains absent. 

Therefore, we comprehensively investigate current environment scaling methods and propose a unified taxonomy aligned with the stages of the \textit{GEF loop}, adopting a pioneering environment-centric perspective. 
In the \textbf{task generation} stage, we categorize methods into \textit{complexity scaling}, \textit{dynamic scaling}, and \textit{diversity scaling}, which together characterize an environment's ability to generate challenging, adaptive, and diverse tasks continuously. In the \textbf{task execution} stage, we highlight \textit{interactivity} and \textit{realism}, since these properties determine the richness and fidelity of the interaction data from which agents learn. In the \textbf{feedback} stage, we categorize the scaling of evaluative signals along \textit{density}, \textit{granularity}, \textit{automation}, \textit{objectivity}, and \textit{robustness}. Beyond this taxonomy, we also analyze implementation frameworks, applications, challenges, and future frontiers. Representative works are listed in Figure~\ref{fig:scaling-environments-survey}. 

The survey is organized as follows. We first categorize representative environment scaling methods along the three-stage taxonomy: task generation (\S\ref{sec:task_gen}), task execution (\S\ref{sec:task_exe}), and feedback (\S\ref{sec:feedback}), highlighting challenges and frontiers in each dimension. We then analyze implementation frameworks in \S\ref{sec:imple} and future directions in \S\ref{sec:future_directions}. More background (\S\ref{sec:background}), and discussions of method comparisons (\ref{sec:compare}), evaluation benchmarks (\S\ref{sec:eval-bench}), and applications (\S\ref{sec:app}) can be found in the Appendix.

\section{Stage 1: Task Generation}
\label{sec:task_gen}
In the task generation stage, the environment is required to propose challenging tasks that push the agent toward its capability boundary. Scaling at this stage targets three aspects of the task design: increasing difficulty\footnote{For clarity, in \textit{complexity scaling}, we only consider the intrinsic difficulty of a task (i.e., static complexity). We group the temporal evolution of task difficulty (dynamic complexity) together with changes in the environment itself under the \textit{dynamic scaling} subsection.} (\textit{complexity scaling} \S~\ref{ssec:task-complexity}), introducing dynamics (\textit{dynamic scaling} \S~\ref{ssec:task-dynamic}), and expanding diversity (\textit{diversity scaling} \S~\ref{ssec:task-diversity}). An illustrative example is shown in Figure~\ref{fig:task}. 

\subsection{Complexity Scaling}
\label{ssec:task-complexity}
Static complexity increases a task's inherent structural intricacy, moving beyond single-step commands to challenges defined by dependencies, logical flows, and hierarchical relationships. A typical scenario is tool use, from early single-step tasks to multi-turn, multi-step scenarios, where complexity scales up as the number of turns and steps increases~\citep{qin2023toolllm, patil2025bfcl, yao2024tau, ye2025toolhop}. More sophisticated tasks exhibit hierarchical or compositional structure, where high-level objectives are decomposed into nested sub-goals, requiring agents to solve novel problems by recombining acquired skills~\citep{shao2023compositional,yu2025c,shi2025taskcraft,wu2025webwalkerbenchmarkingllmsweb,wu2025webdancerautonomousinformationseeking}. Building on this, TaskCraft~\citep{shi2025taskcraft} expands tasks' structure both in depth (longer sequences of tool executions) and in width (multiple sub-goals per objective), thus enhancing hierarchical reasoning. Further escalating this complexity, conditional and graph-based tasks involve non-linear structures with branching logic, where planning must adapt dynamically to intermediate outcomes~\citep{ tao2025webshaperagenticallydatasynthesizing, li2025websailornavigatingsuperhumanreasoning,tao2025webleaper,liu2025webexplorer}. Such designs lead to significant performance gains, as shown in Table \ref{tab:complexity}, highlighting the importance of scaling up tasks' complexity to foster robust and generalizable reasoning strategies.

\paragraph{\emojienvisage~Future Frontiers}
The next step in scaling up task complexity involves: 1) broadening the width by expanding from a limited set of tools to more versatile, multimodal tools, and addressing increasingly complex scenarios that go beyond simple tool invocation and web navigation, and 2) increasing the depth. However, most current errors arise due to context limitations, and synchronous RL frameworks are inefficient at supporting the scaling of task depth, which further exacerbates these issues. Future environmental designs should prioritize supporting flexible, independent updates through asynchronous RL frameworks.

\subsection{Dynamic Scaling}
\label{ssec:task-dynamic}
Scaling dynamics focuses on creating an environment that changes task complexity based on agent performance, helping agents learn and adapt to new challenges and goals. This involves scheduling tasks of varying difficulty levels (curriculum learning) based on agent performance metrics, such as success rate (SR) or progress rate (PR), or targeting newly acquired or weaker skills for focused training~\citep{zeng2025rlve,liang2024environment,zala2024envgen,qiwebrl,xi2024agentgym,cheng2025evocurr,zala2024envgen}. For instance, AgentGen~\citep{hu2025agentgen} introduces a bidirectional task variation method, adjusting task complexity upwards or downwards to match the agent's capabilities, while EnvGen~\citep{zala2024envgen} iteratively generates embodied environments that focus on the skills where agents are weak at. More recent advances introduce a challenger-solver co-evolving pipeline, where the challenger proposes near-boundary tasks based on the solver's uncertainty, and the solver improves by training on filtered task sets, resulting in progressively more difficult curricula and improved reasoning performance~\citep{huang2025r}. A similar conclusion is drawn by \citet{zeng2025rlve}, where adaptively shifting a sliding window of difficulty (ensuring appropriately challenging problems throughout training) leads to superior performance gains and improved learning efficiency.

\paragraph{\emojienvisage~Future Frontiers}
Despite the benefits of adaptively changing environments and task difficulties, most current methods are limited to verifiable tasks or require manual intervention to ensure high-quality task generation~\citep{huang2025r,zeng2025rlve}. Future designs may focus on developing systematic principles for non-verifiable tasks while also making automated task generation more reliable.

\subsection{Diversity Scaling}
\label{ssec:task-diversity}
Scaling the diversity of environments and tasks is crucial for preventing agents from overfitting to specific action patterns, thereby fostering more robust and generalizable performances \citep{hu2025agentgen, huang2025r}. Although \textit{task-level diversity scaling} methods can generate tasks with different objectives and levels of complexity within a fixed environment \citep{huang2025r}, enabling the agent to gradually improve, \citet{andrews2025arescalingagentenvironments} argues that simple tasks in complex environments may be more effective than complex tasks in simple ones. This suggests that greater emphasis should be placed on \textit{environment-level diversity scaling}. For instance, AgentGym~\citep{xi2024agentgym} and AgentBank~\citep{song2024agentbank} utilize environments from various domains (e.g., Web, embodied AI, tool use, and programming) showing that heterogeneity improves performance on both in-domain and out-of-domain tasks. More recent advances explore to automatically construct agentic environments from either pre-defined \citep{shi2025taskcraft,andrews2025arescalingagentenvironments,yang2025swesmithscalingdatasoftware,jain2025r2egymproceduralenvironmentshybrid} or continuously refined \citep{hu2025agentgen,zhang2025autoenv} tool and application suites. As shown by \citet{zhang2025autoenv}, although no single learning strategy can handle all cases as environment diversity scales up, the performance upper bound achievable across all environments increases consistently.

\paragraph{\emojienvisage~Future Frontiers} A key challenge for scaling diversity is the diminishing marginal utility of merely adding more heterogeneous environments. Future environment designs should therefore not only scale diversity, but develop adaptive mechanisms for selecting or composing learning strategies tailored to heterogeneous environments \citep{zhang2025autoenv}.

\section{Stage 2: Task Execution}
\label{sec:task_exe}

In the task execution stage, interaction takes the form of the agent taking an action and receiving an observation from the environment. We organize works towards environment scaling in this stage into two dimensions: interactivity scaling and realism scaling (Figure~\ref{fig:task}).  \textit{Interactivity} (\S~\ref{ssec:task-interactivity}) learns how agent performance affected by the frequency of tool interaction, while \textit{realism} (\S~\ref{ssec:task-realism}) focuses on making observations more realistic to enhance the interaction quality.

\subsection{Interactivity Scaling}
\label{ssec:task-interactivity}
Interactivity concerns whether the environment is executable and whether the agent is blind to the returns of intermediate steps. In non-interactive settings, agents are evaluated against a single static ground-truth solution path, even though there may in fact exist diverse paths to successful task completion \citep{deng2023mind2web,kapoor2024omniact,rawles2023androidwild, qin2023toolllm}. Consequently, such static supervision fails to credit alternative correct solutions and limits agents' ability to generalize to novel tasks \citep{sullivan2025procedural,xie2024osworld,rawles2025androidworld}. Recent environment designs start to allow agents to interactively invoke real-world APIs, call functions, generate codes, or interact through GUI \citep{wang2024executable, yao2023webshopscalablerealworldweb, trivedi2024appworldcontrollableworldapps, sullivan2025procedural} and to dynamically adapt subsequent actions based on intermediate results. \textit{Width scaling methods} focus on enabling parallel tool invocation \citep{pang2025browsemaster} and broadening the interaction dimension (e.g., facilitating flexible switching between GUI operations and tool invocations \citep{jia2025osworldmcp}), while \textit{depth scaling methods} focus on efficient tool-use context management. For example,  MiroThinker \citep{miromindteam2025mirothinker} removes weakly related tool calls from the context, thereby freeing up capacity to scale up tool-use depth and yielding consistent performance gains across various benchmarks.

A bottleneck for such scaling lies in the cost of APIs calls. To alleviate this issue, several recent advances implement an offline realistic database as the environment \citep{tongyidr, fang2025generalagentic, ye2025feedbackdriven}, using function calls to read from and write to the database state, thereby keeping the cost of scaling up interactivity within a reasonable budget while preserving some realism.

\paragraph{\emojienvisage~Future Frontiers} While scaling interactivity enables richer tool use, it also introduces problems like overlong responses, diminishing marginal returns from additional tool calls, and escalating costs \citep{miromindteam2025mirothinker}. Moreover, efficacy diminishes in scenarios requiring the combination of multiple tools \citep{jia2025osworldmcp}, especially as the emerging paradigm of thinking with images \citep{su2025thinking} introduces multi-modal tool-call returns, making their fusion significantly more challenging. Future environment designs should therefore prioritize optimizing the efficiency and quality of heterogeneous real-world tool invocation.

\subsection{Realism Scaling}
\label{ssec:task-realism}
Realism aims to ensure that observations obtained from the environment remain consistent with the real world. \textbf{In tool-use scenarios}, despite the benefits brought by the Model Context Protocol (MCP) \citep{anthropic2024mcp, luo2025mcp, wang2025mcp, fan2025mcptoolbench++}, which integrates heterogeneous tools into a unified context and greatly improves the efficiency of tool use, directly calling APIs in real environments still faces challenges of monetary cost and lack of robustness \citep{song2023restgpt, wu2025webwalkerbenchmarkingllmsweb, mastouri2025makingrestapisagentready,prabhakar2025apigenmt}. Previous \textit{model-based} environment designs use LLMs as a world model \citep{hao2023reasoninglanguagemodelplanning,ge2024worldgpt,tang2024worldcoder} or to simulate the results of tool calls \citep{qin2023toolllm, lu2025toolsandbox, sun2025zerosearch}, reducing overhead and avoiding implementation errors but inevitably sacrificing realism. Facing this dilemma, \textit{offline-database methods} \citep{tongyidr, fang2025generalagentic, ye2025feedbackdriven} make a trade-off by using a snapshot of a real-world database as the interaction environment, thereby preserving a relatively realistic simulation (e.g., tool-chain failures, asynchronous external events). \textbf{In Computer-Use scenarios,} earlier works mainly construct on Q\&A over static webpage screenshots \citep{mialon2023gaia,deng2023mind2web,sun2022metagui,kapoor2024omniact}, rather than actually operating in real executable environments. This idealized setting makes agents brittle and prone to failure in real-world conditions (e.g. similar buttons, intrusive popup windows, and heterogeneous user interfaces) that demand context awareness and fine-grained control \citep{xie2025scalingcomputerusegroundinguser}. To mitigate these, recent advances instead adopt real executable environments that involve coding \citep{yang2023intercodestandardizingbenchmarkinginteractive}, web browser navigation \citep{yao2023webshopscalablerealworldweb,zhou2023webarena}, and application control across operating systems \citep{xie2024osworld}.

When extending to the \textbf{multi-agent setting}, interactions are no longer limited to those between agents and tools, but also occur between agents themselves. Agents may coordinate or compete with one another, so each agent’s behavior naturally becomes part of the environment for the others \citep{tran2025multi, qian2024scaling, li2024smoa, zhang2024chain, barres2025tau2bench}. Under this setting, realism scaling concerns whether the environment provides reliable underlying infrastructure and communication mechanisms \citep{andrews2025arescalingagentenvironments, zhang2025parallelized, li2023camel, gao2024agentscope} such that, as the number of agents grows, complex social and economic phenomena (e.g., information diffusion, opinion polarization, and herding effects \citep{yang2025oasis, zhang2025parallelized}) can emerge and be meaningfully simulated. For instance, \citet{zhang2025parallelized} adopts the MQTT communication protocol to support asynchronous decision making among autonomous agents,  while \citet{andrews2025arescalingagentenvironments} decouples agent and environment clocks and treats other agents’ activities as independent events, thereby allowing the world state to evolve asynchronously and supporting the collection of more realistic experiential data.

\paragraph{\emojienvisage~Future Frontiers} Realistic simulation has long been a fundamental open challenge. Most existing environments follow a ReAct-style paradigm  \citep{yao2023react}, assuming that the environment remains static between an agent’s observation and action. However, information arrives as overlapping streams in real world so that perception and action should occur concurrently \citep{andrews2025arescalingagentenvironments}. Future environment designs should place greater emphasis on such asynchronous settings, while also exploring ways to unlock richer scenarios and modalities.


\begin{figure*}[t]
\includegraphics[width=\textwidth]{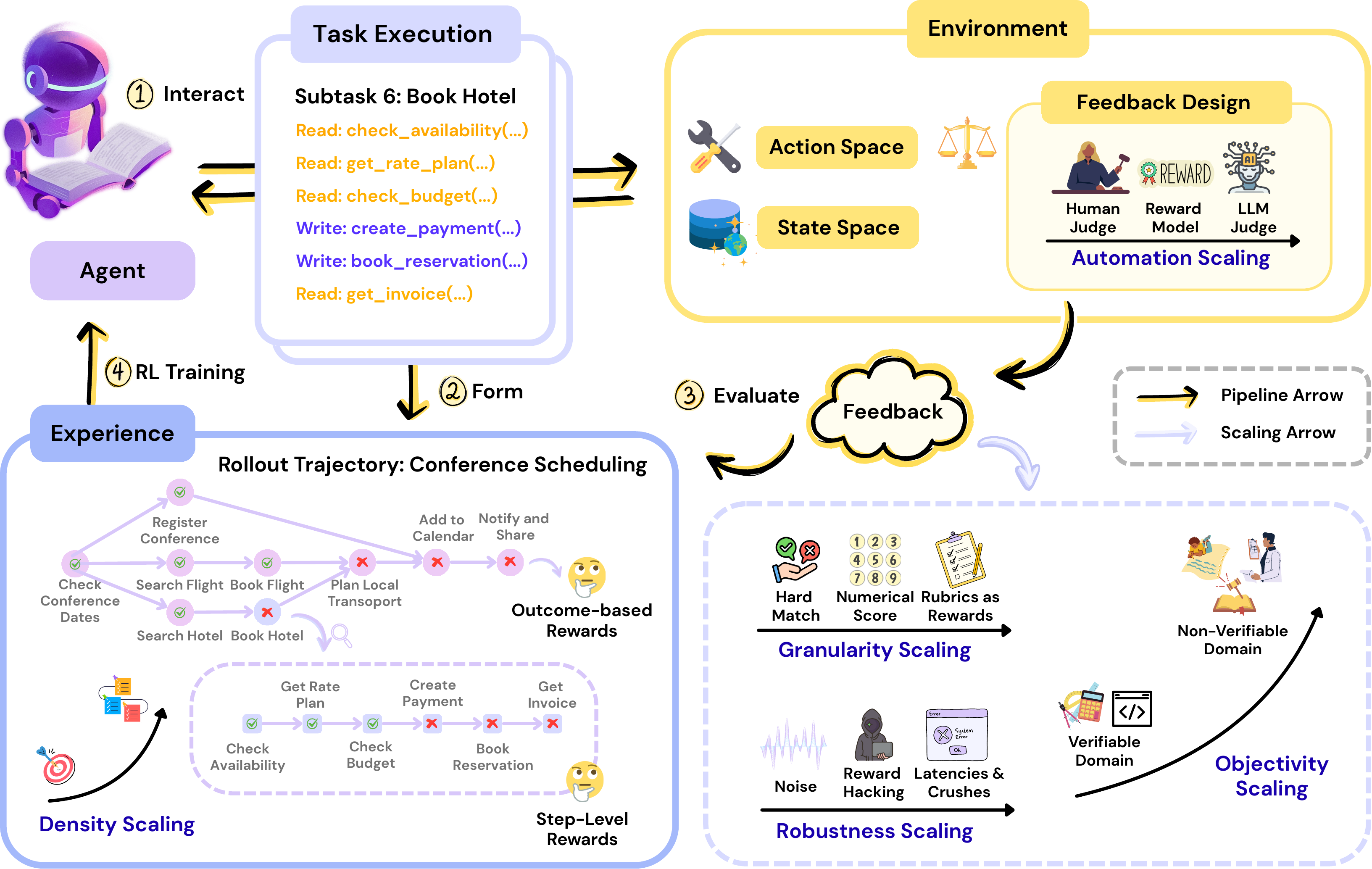}
\vspace{-0.6em}
\caption{Illustration of environment scaling in the \textbf{feedback} stage using a conference-scheduling example. The agent first executes tasks in the environment and produces action-observation trajectories. The environment then evaluates these trajectories and returns feedback, yielding the experience used to train the agent. Scaling in the feedback stage covers \textit{density}, \textit{granularity}, \textit{automation}, \textit{objectivity}, and \textit{robustness}.}
\label{fig:feedback}
\end{figure*}

\section{Stage 3: Feedback}
\label{sec:feedback}
In the feedback stage, the environment assesses the trajectories collected during task execution and provides feedback for subsequent training. Scaling at this stage focuses on how feedback is provided, including its frequency and richness (\textit{density} \S~\ref{ssec:feedback-density} and \textit{granularity} \S~\ref{ssec:feedback-granularity}), its level of automation (\textit{automation} \S~\ref{ssec:feedback-automation}), as well as how objectively and reliably it is delivered (\textit{objectivity} \S~\ref{ssec:feedback-objectivity} and \textit{robustness} \S~\ref{ssec:feedback-robustness}), as shown in Figure~\ref{fig:feedback}. 

\subsection{Density Scaling}
\label{ssec:feedback-density}
The density of feedback refers to how frequently evaluative signals are provided, typically including trajectory-level outcome-based rewards and step-level process-based rewards. \textit{Trajectory-level outcome-based rewards}~\citep{jiang2025verltool, da2025agent, qian2025toolrl, wang2025otc} are relatively sparse and contribute to more stable training, especially in tasks like mathematical reasoning and code generation~\citep{deepseekai2025deepseekr1incentivizingreasoningcapability}. However, it is challenging to credit whether an intermediate step contributes to or hinders task completion in more complex, multi-step tasks. \textit{Step-level process-based rewards}~\citep{khalifa2025processrewardmodelsthink,chae2025webshepherd,gao2024designing,ma2025steplevelrewardmodelsrewarding,park2025know}, often combined with outcome-based rewards, in contrast, provide denser supervision and more detailed guidance. Still, their benefits are not consistent, with performance in natural language tasks (e.g., creative writing) being inferior to that in logical-coherent tasks (e.g., coding), and they face the problem of reward hacking, where trivial patterns receive high rewards~\citep{ma2025steplevelrewardmodelsrewarding,gao2024designing}. Efforts like PR-Clip-Delta~\citep{gao2024designing} utilize reward differences between adjacent steps as rewards, clipping them if they exceed a predefined threshold, which helps mitigate repetitive behavior and the training instability associated with step-level rewards.

\paragraph{\emojienvisage~Future Frontiers} 
Future reward design needs to further explore the underlying mechanisms of reward models as well as their interpretability. Efforts should aim to combine the training stability of trajectory-level rewards and the fine-grained evaluation benefits of step-level rewards, expanding their application to more domains. Meanwhile, as current reward models are mostly poorly calibrated~\citep{park2025know}, assigning overly optimistic scores to some intermediate steps, future designs should focus on addressing such biases.

\subsection{Granularity Scaling}\label{ssec:feedback-granularity}
Granularity scaling refers to increasing the level of detail in feedback as well as enriching the forms of feedback provided. In earlier stages, evaluative feedback typically consisted of binary signals \citep{NIPS2017_d5e2c0ad, ibarz2018rewardlearninghumanpreferences} or a single numeric score \citep{stiennon2022learningsummarizehumanfeedback, ouyang2022traininglanguagemodelsfollow}. At finer granularities, the evaluative feedback on agents' performances is broken down into structured components, such as a set of scores across multiple criteria like correctness and format, offering more informative guidance~\citep{huang2025adactrl,qian2025toolrl, song2025r1}. The \textit{Rubrics as Rewards} framework~\citep{gunjal2025rubricsrewardsreinforcementlearning,huang2025reinforcement,zhou2025breaking,zhang2025chasingtail,viswanathan2025checklist} further decomposes task requirements into tangible, human-interpretable criteria. By designing rewards as checklist-style, instance-specific rubrics, it provides a middle ground between binary correctness signals and broad preference rankings.

\paragraph{\emojienvisage~Future Frontiers} 
Despite rubrics at finer granularity providing more detailed information than a simple overall numeric evaluation, their predefined nature by humans and reliance on internal, maybe outdated parametric knowledge will lead agents to exploit internal patterns rather than explore external knowledge~\citep{shao2025drtulu}. Future work should allow rubrics to co-evolve with the expanding boundaries of knowledge.

\subsection{Automation Scaling}
\label{ssec:feedback-automation}
Automation scaling involves the shift of the feedback mechanism from slow, costly, and labor-intensive human feedback to automated evaluations. The previous \textit{Reinforcement Learning from Human Feedback (RLHF)} framework heavily relies on human annotators, making it difficult to scale~\citep{Sheng_2025, hu2025openrlhfeasytousescalablehighperformance, christiano2017deep}. Recent works adopt LLM-as-a-Judge, train generative reward models, or utilize agentic workflows with external tools to facilitate the feedback process. \textit{LLM-as-a-Judge}~\citep{chen2025judgelrmlargereasoningmodels, zheng2023judgingllmasajudge,gunjal2025rubricsrewardsreinforcementlearning,huang2025reinforcement} directly prompts off-the-shelf LLMs to provide comparative preferences or numeric scores across multiple aspects. \textit{Training-based methods}~\citep{chen2025scalingautonomousagentsautomatic,zhang2025generativeverifiersrewardmodeling} train a reward model with synthesized rationales or trajectories to classify whether user intent is fulfilled, significantly outperforming simple prompt engineering but at the cost of limited generalization and higher training costs. \textit{Agentic methods}~\citep{peng2025agenticrewardmodelingintegrating} further incorporate external searching tools to help verify factual accuracy and assess instruction-following ability, offering more reliable guidance, though there is still room for improvement in workflow efficiency and reward accuracy.

\paragraph{\emojienvisage~Future Frontiers} 
Automation magnifies risks such as the propagation of biases inherent in LLMs, like verbosity bias (favoring longer responses), position bias (favoring answers in certain positions), egocentric bias (preferring self-generated responses), and so on~\citep{zheng2023judgingllmasajudge}. It also introduces the risk of reward hacking, where intelligent agents exploit loopholes in the reward model to achieve high scores, thereby deviating from the intended goals and even causing the model to crash. Future work should focus on balancing the critical trade-off between scalability and security, with an emphasis on developing robust risk mitigation strategies as automation scales.

\subsection{Objectivity Scaling}
\label{ssec:feedback-objectivity} 
Automated verification by LLMs may inevitably introduce biases and noise, while human evaluation is also subjective and limited by constrained knowledge. Therefore, it is necessary to eliminate subjectivity in environmental feedback designs. Objectivity scaling aims to reduce these biases and make automated verifiers fairer. Despite the \textit{Reinforcement Learning with Verifiable Rewards (RLVR)} paradigm~\citep{lambert2025tulu3} achieving great success in some easy-to-verify domains (e.g., mathematical reasoning, code generation~\citep{shao2024deepseekmathpushinglimitsmathematical, deepseekai2025deepseekr1incentivizingreasoningcapability}), it still falls short in creative domains (e.g., writing, medical consultation, and policy making), where ground truth is often missing, and verification remains open, making it difficult to reach a unified standard. Recent advances aim to extend RLVR to these hard-to-verify domains~\citep{andrews2025arescalingagentenvironments, ma2025general, akter2025nemotron, su2025crossingrewardbridgeexpanding, yu2025rlpr, liu2025nover, gunjal2025rubricsrewardsreinforcementlearning, huang2025reinforcement,li2025omni}. For instance, BRPO~\citep{jia2025writingzerobridgegapnonverifiable} introduces a pairwise generative reward model designed to offer comparative quality scores, avoiding the direct scoring of creativity in writing materials. Works like ARE~\citep{andrews2025arescalingagentenvironments} combine parameter-matching hard checks (e.g., verifying email id) with generative rubric-based soft checks to ensure relatively fair evaluation.

\paragraph{\emojienvisage~Future Frontiers} 
It is struggling to provide a unified standard during evaluation in domains like creative writing. In future work, when extending to broader domains and more modalities (e.g., the aesthetic evaluation of images and videos), the challenges of subjectivity will become even more pronounced. Extending RLVR to more scenarios or finding other ways to eliminate subjectivity will remain a crucial open challenge and a key frontier in the development of more intelligent agents.

\subsection{Robustness Scaling}
\label{ssec:feedback-robustness}
Feedback robustness requires the environment to provide stable and reliable reward signals. \textit{Reward-level robustness} concerns the stability of the reward itself, since such signals may be inherently noisy or susceptible to hacking (agents learning undesirable and tricky behavioral patterns that obtain high rewards without achieving the intended goals). To mitigate noise, \citet{lin2024navigating} and \citet{su2025crossingrewardbridgeexpanding} generate soft probabilistic rewards with generative verifiers. To prevent reward hacking, \citet{farquhar2025monamyopicoptimizationnonmyopic} evaluates the future utility of actions through an overseer, constraining unstable behaviors while preserving explainability. \citet{huang2025reinforcement} develops a defense rubric that penalizes sycophantic praise toward user prompts and overly flattering self-assessments in responses, encouraging the model to produce more substantive content. \textit{Environment-level robustness} is also crucial for agents, as environmental instabilities (e.g., delays, crashes, corrupted tool outputs) may also undermine feedback reliability and degrade the training process. To handle such failures, Trinity-RFT \citep{pan2025trinityrftgeneralpurposeunifiedframework} proposes asynchronous inference and retry mechanisms, while \citet{tongyidr} employs caching, retrying failed calls, and switching to similar providers to prevent corrupted trajectories. Both approaches contribute underlying frameworks that ensure stable feedback.

\paragraph{\emojienvisage~Future Frontiers} Future reward designs should prioritize both efficacy and the prevention of hacking through tricky patterns, with greater attention also given to the development of more robust feedback mechanisms and underlying frameworks.

\section{Implementation}
\label{sec:imple}
\subsection{Environmental Designs}

In this section, we discuss environmental design choices across domains. In \textbf{Tool-Use} scenarios, environmental designs prioritize expanding the interactivity width and depth. This involves supporting deeper tool call sequences to solve more complex tasks and incorporating more general, versatile tools to handle tasks across a broader range of domains. 

Similarly, \textbf{Deep Research} scenarios focus on enabling agents to orchestrate diverse search and browsing APIs during task generation, as well as robust retrieval and reasoning through long contexts. For example, DR Tulu agent infrastructure~\citep{shao2025drtulu} optimizes a unified MCP-based tool backend through global caching and asynchronous process lock, supporting high-concurrency operations. ARE~\citep{andrews2025arescalingagentenvironments} further transforms tool-use benchmarks into separate applications, integrating mobile use and tool use into a unified environment. 

In \textbf{Software Engineering (Coding)} scenarios, designs emphasize whether the agent can actually execute code and whether the process can be reproduced. Docker~\citep{merkel2014docker} is commonly utilized as the basis of execution sandbox, as it ensures consistency across different machines through standard Dockerfiles while also enabling the safe execution of code within an isolated virtual environment, preventing potential safety issues~\citep{yang2023intercodestandardizingbenchmarkinginteractive,jimenez2023swe,yang2025swesmithscalingdatasoftware}. 

\textbf{Web Navigation} scenarios also focus on realism and interactivity scaling. WebArena~\citep{zhou2023webarena} delivers websites in self-contained Docker images as well, avoiding unexpected content or configuration changes. WebShop~\citep{yao2023webshopscalablerealworldweb} offers a diverse range of semantic action space (searching queries and choosing buttons), with its modular architecture facilitating scaling to new web tasks and domains. VisualWebArena~\citep{koh2024visualwebarena} further introduces a novel Classifieds website built on OSClass, a robust open-source Content Management System (CMS), supporting visually grounded tasks such as posting, searching, and commenting, thereby offering a more interactive environment for user engagement. 

In \textbf{Computer Use} scenarios, in addition to interactivity and realism, data collection efficiency and diversity integration (e.g., cross-OS tasks, heterogeneous tools) are also key concerns. OpenCUA~\citep{wang2025opencua} and OSWorld~\citep{xie2024osworld} provide infrastructures that support efficient task annotation and the real-time capture of user interactions. UI-TARS-2~\citep{wang2025uitars2} unifies GUI operations and SDK function calls into a versatile platform, supporting a variety of task kinds and maintaining long-lived states across intricate, multi-step interactions. 

In \textbf{Embodied} scenarios, tasks involve requiring agents to follow instructions, ground objects in photo-realistic 3D spaces, answer questions, or perform multi-step tasks~\citep{weihs2021visualroomrearrangement,das2018embodied}. These tasks require environments to maintain real-world realism, such as actual grounding and collision avoidance. Frameworks like ALFWorld~\citep{shridhar2020alfworld} provide the necessary infrastructure for abstract reasoning and concrete execution. More recently, Genie 3~\citep{parker2025genie3} has further improved real-time interactivity and long-horizon consistency, providing a fertile foundation for the advancement of future LLM agents.

\subsection{Implementation Challenges}
A key challenge in environmental design is the \textit{Generator-Verifier Asymmetry}~\citep{wei2025asymmetry}, which refers to the mismatch between the intelligence required for the generator and that required for the verifier. In \textit{easy-to-verify} domains (e.g., mathematics and programming~\citep{wei2025browsecomp, jimenez2023swe, phan2025humanity}), task generation can be difficult (scaling up high-quality, non-trivial tasks is challenging), while verification is relatively easy and computationally inexpensive. In contrast, in \textit{easy-to-generate} domains (e.g., creative writing and healthcare~\citep{lin2024wildbench, arora2025healthbench}), it is easy to propose open-ended and challenging tasks, but verification may involve subjectivity or require expert knowledge. Future environmental designs may allow the generator and verifier to co-evolve, tapping into the potential of using a powerful generator as a verifier to stimulate the development of even more powerful generators~\citep{huang2025r, hong2025cooper, lu2025urpo, chen2025self, wang2025socraticzerobootstrappingreasoning}. See more details are in Appendix \ref{sec:background}.

\section{Future Directions}
\label{sec:future_directions}
\paragraph{Co-Evolution via Embedded External Tools}
As the complexity of task increases, feedback mechanisms must scale in order to avoid sparse, noisy, or misaligned learning signals. Future environments can achieve this by including external tools or modules to serve as verifiers, simulators, compilers, or executable systems into the learning loop. These tools would evolve together with the task generation, providing structured, verifiable feedback and enabling agents to interact with increasingly sophisticated challenges. By integrating embedded tools with automated evaluators such as LLM-based formative signals, environments can better support multi-step, open-ended, and creative problem solving, while also helping to mitigate the Generator-Verifier Asymmetry.

\paragraph{Scaling Through Generator-Verifier Synergy} 
Future environments can encourage stronger generators to have the ability to decompose complex tasks into smaller subproblems with intermediate solutions, making them tractable for weaker verifiers. This enables scalable supervision in domains where holistic verification is difficult such as creative cultural production and policy-making. In contrast, weaker generators could provide diverse candidate solutions that could be filtered, ranked, or refined by stronger verifiers. By incorporating these co-evolving dynamics into the environment, the asymmetry between generators and verifiers can serve as a catalyst for continuous self-improvement.

\paragraph{Open-Ended, Multi-Agent Environments}
Future environments can scale to support large-scale multi-agent interactions, emergent social dynamics, and economic or organizational level simulations, providing rich contexts for studying both collaborative and non-cooperative behaviors in complex environments. In particular, scaling to massively multi-cultural and multi-lingual settings requires scaling environment construction for agents to navigate the subtle semantics of concepts and values that vary across different societies, not only in the textual but also in the multi-modal domain \citep{koh-etal-2024-visualwebarena,huangcultureclip}. Such open-ended and interactive scenarios environments foster generalization and strategic planning abilities of agents, equipping LLM agents to better handle complex, real-world challenges beyond isolated task execution.

\section{Conclusion}
\label{sec:conclusion}
The era of experience makes environments central to the development of LLM agents, positioning them as active producers of experiential data and underscoring the growing need for scaling environments to create a more complex, realistic, and richly interactive world. From a pioneering environment-centric perspective, our survey proposes a unified taxonomy that organizes representative work across the three-stage \textit{GEF loop} (task generation, task execution, and feedback), together with implementation frameworks and key applications. Additionally, we surface key challenges and highlight future frontiers for advancing agent intelligence along each scaling dimension, offering valuable insights for future research on agentic systems.

\section*{Limitations}

As scaling environments remains an emerging research topic, relatively few studies have explicitly adopted this framing. Thus, we take a broad view of environment and organize representative studies along the \textit{GEF loop} (task generation, task execution, and feedback) from a pioneering environment-centric perspective. While this broader lens brings in some adjacent lines of work that may not have been explicitly designed from the environment side, our taxonomy is both comprehensive and insightful. Given the rapid pace of agentic research, some of the most recent papers may fall outside this snapshot. We will continue to update this survey as the literature evolves.

\section*{Acknowledgments}
We thank the anonymous reviewers for their valuable suggestions and comments. Yuchen Huang is supported by the Hong Kong Ph.D. Fellowship Scheme (HKPFS).

\bibliography{custom}

\appendix

\section{Background}
\label{sec:background}
\subsection{Scaling Laws for LLM Agents} 
Just as large language models exhibit predictable performance scaling with increases in the number of parameters, the volume of training data, and the compute budget, agent systems likewise display scaling regularities along three axes: (i) expanding the agent population and identifying properties that emerge as interactions increase; (ii) increasing environmental complexity and assessing how realistic, dynamic settings shape learning and adaptation; and (iii) extending the horizons of evolution and memory to study how agents generalize and improve through accumulated experience \citep{camel2025scaling}. While most existing surveys on LLM agents adopt an agent-centric view \citep{luo2025largelanguagemodelagent,xi2025rise,yehudai2025survey,gao2025survey}, covering topics from multi-agent interaction \citep{qian2024scaling,tran2025multi} to self-evolution \citep{gao2025survey,tao2024surveyselfevolutionlargelanguage}, environment scaling remains underexplored and has not been systematically organized. In this work, we take an environment-centric perspective on scaling environments and examine how dynamic, richly interactive, high-fidelity worlds can accelerate agent development and evolution.

\subsection{Generator-Verifier Asymmetry}
\label{ssec:asymmetry}

\begin{figure}[b]
\centering
\includegraphics[width=0.8\columnwidth]{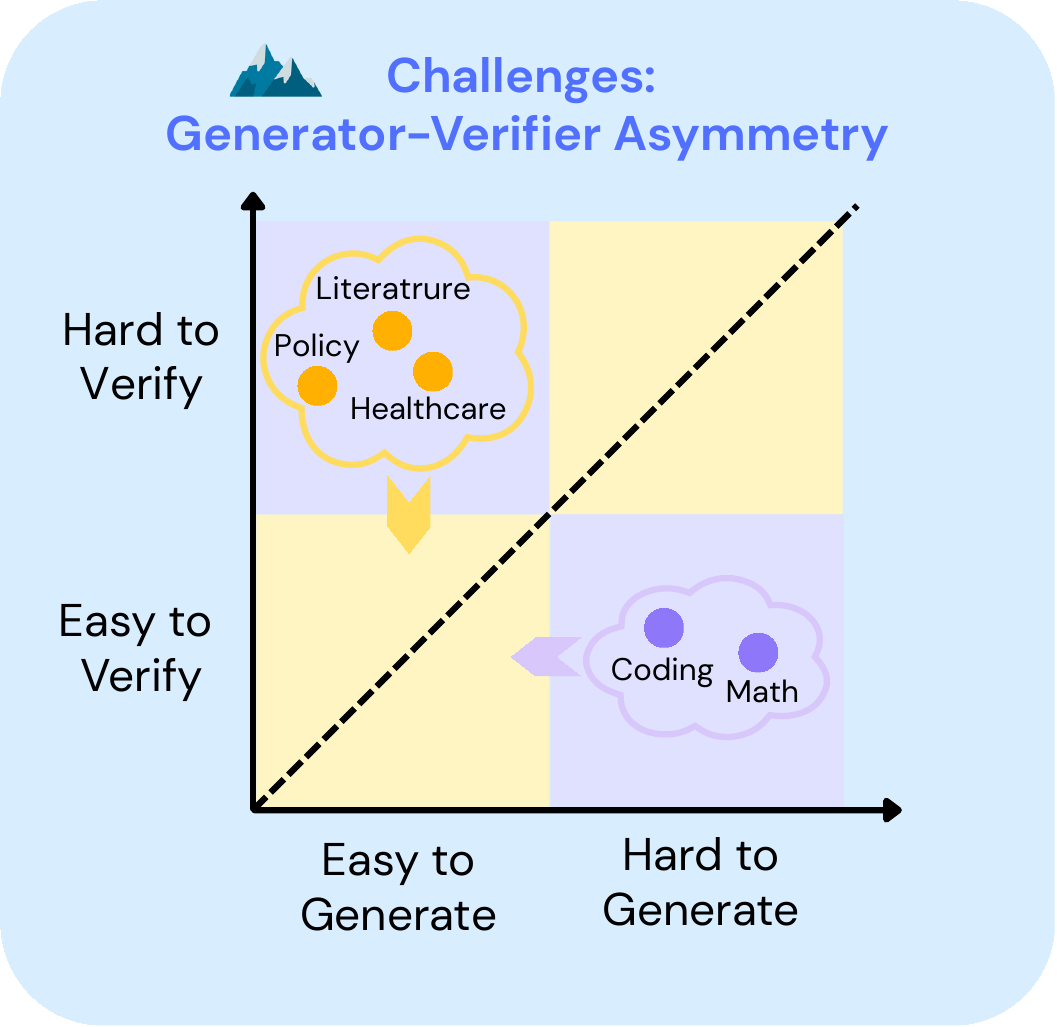}
\vspace{-0.6em}
\caption{Generator-verifier asymmetry challenge.}
\label{fig:asymmetry}
\end{figure}

A fundamental characteristic in many real-world tasks is the inherent \emph{Generator-Verifier Asymmetry}~\citep{wei2025asymmetry}, namely the mismatch between the intelligence required for generator, which generates (\S\ref{sec:task_gen}) or executes (\S\ref{sec:task_exe}) tasks, and that required for verifier, which provides feedback (\S\ref{sec:feedback}). These two kinds of intelligence naturally form two axes critical to next-generation Agentic AI, as illustrated in Figure~\ref{fig:asymmetry}. From this perspective, scaling up environments essentially corresponds to scaling intelligence along the $x$-axis and the $y$-axis. Current progress in RL largely exploits the regime on the easy-to-verify side of this asymmetry. These \textbf{Easy-to-Verify, Hard-to-Generate Domains} include fields such as mathematics and programming~\citep{wei2025browsecomp, jimenez2023swe, phan2025humanity}. For these domains, generating and solving a continual stream of high-quality, non-trivial tasks is challenging. In contrast, verification is objective and computationally inexpensive (e.g., via unit tests or exact match on mathematical results). This enables weak-to-strong supervision, where a simple verifier can provide accurate feedback to train a much stronger agent for solving hard tasks. On the contrary, the \textbf{Hard-to-Verify, Easy-to-Generate Domains} include areas such as creative writing, policy-making, or healthcare~\citep{lin2024wildbench, arora2025healthbench}. For these easy-to-propose, open-ended tasks, verification is subjective, requires substantial expert judgment, or unfolds over long horizons, making high-quality feedback scarce and expensive. This bottleneck, corresponding to the upper-left region of coordinate system, poses more difficulty in modeling the environment, and rendering environment scaling more challenging yet offering greater potential for advancing agent capabilities. Notably, the asymmetry also presents an opportunity: if the generator's stronger intelligence can be systematically leveraged to strengthen the verifier, so that it can supervise an even stronger generator, then such asymmetric property can be exploited to drive agents' self-evolution~\citep{huang2025r, hong2025cooper, lu2025urpo, chen2025self, wang2025socraticzerobootstrappingreasoning}.

\subsection{Conceptual Framework}
Following the formalization by \citet{gao2025survey}, we model the environment $E$ as a partially observable Markov decision process (POMDP). At the beginning of each episode, the environment generates a task $T = (E, I)$, where $I \in \mathcal{I}$ represents user intention drawn from intention space $\mathcal{I}$. The agent $\pi$ interacts with the environment over horizon $T \in \mathbb{N}$, producing an interleaved observation--action trajectory $\tau = (o_0, a_0, o_1, a_1, \ldots, o_T) \in (\mathcal{O} \times \mathcal{A})^{T+1}$, where $\mathcal{O}$ and $\mathcal{A}$ denote the observation and action spaces respectively. The environment then evaluates performance and provides feedback $r \in \mathbb{R}^k$, which may take the form of step-level signals $r_{\text{step}}^{(t)} = R_{\text{step}}(s_t, a_t, T) \in \mathbb{R}$ for $t \in \{0, \ldots, T\}$, where $s_t \in \mathcal{S}$ denotes the state of the environment at step $t$, trajectory-level signals $r_{\text{traj}} = R_{\text{traj}}(\tau, T) \in \mathbb{R}^m$, or a combination of both $r = f(r_{\text{step}}^{(0:T)}, r_{\text{traj}})$ where $f : \mathbb{R}^{T+1} \times \mathbb{R}^m \to \mathbb{R}^k$. These signals need not be limited to sparse scalar rewards but can encode structured or adaptive assessments reflecting correctness, efficiency, reasoning depth, or long-term outcomes. This \textit{Generation-Execution-Feedback (GEF) loop} $\mathcal{L} = (T_{\text{gen}}, \text{Exec}, \text{Eval}) : \mathcal{I} \to \mathcal{T} \times (\mathcal{O} \times \mathcal{A})^* \times \mathbb{R}^k$, which encompasses task generation, task execution, and feedback, defines the essential mechanics of environments. Repeated iterations drive the accumulation of experience and the progressive evolution of the agent $\pi$.

\section{Comparative Analysis}
\label{sec:compare}
A comparative analysis of methods for scaling task complexity and task diversity is presented in Tables \ref{tab:complexity} and \ref{tab:diversity}.

\begin{table*}[h]
\centering
\caption{Comparative Analysis of Task Complexity and Performance Across Methods}
\label{tab:complexity}
\resizebox{\textwidth}{!}{
    \begin{tabular}{lcccccc} 
        \toprule
        \multirow{2}{*}{\textbf{Method}} 
        & \multirow{2}{*}{\textbf{Skill Dimension}} 
        & \multirow{2}{*}{\textbf{Task Structure}} 
        & \multicolumn{2}{c}{\textbf{Task Complexity}} 
        & \multirow{2}{*}{\textbf{Multi-Turn}} 
        & \multirow{2}{*}{\textbf{Performance}} \\
        \cmidrule(lr){4-5}
        & & & \textbf{Width} & \textbf{Depth} & & \\
        \midrule
        ToolBench \citep{qin2023toolllm} & Tool Use & Sequential & 1 & - & $\times$ & - \\
        $\tau$-bench \citep{yao2024tau} & Tool Use & Sequential & 1 & -, 30 & $\checkmark$ & - \\
        BFCL V3 \citep{patil2025bfcl} & Tool Use & Sequential & 1 & -, 20 & $\checkmark$ & - \\
        $c^3$-bench \citep{yu2025c} & Tool Use & Compositional & 1$\sim$4 & - & $\checkmark$ & - \\
        TaskCraft \citep{shi2025taskcraft} & Tool Use & Compositional & 2 & 1$\sim$4, 7 & $\times$ & - \\
        WebRL \citep{qiwebrl} & Web & Compositional & 1$\sim$4 & -, \(>\)10 & $\checkmark$ & - \\
        WebDancer-32B \citep{wu2025webdancerautonomousinformationseeking} & Web & Compositional & - & 2$\sim$4, 10 & $\times$ & 40.7 \\
        WebShaper-32B \citep{tao2025webshaperagenticallydatasynthesizing} & Web & Graph-Based & - & 3$\sim$7, 30 & $\times$ & 52.4 \\
        WebSailor-32B \citep{li2025websailornavigatingsuperhumanreasoning} & Web & Graph-Based & - & 2$\sim$8, 30 & $\times$ & 53.2 \\
        WebExplorer-8B \citep{li2025websailornavigatingsuperhumanreasoning} & Web & Graph-Based & - & 0$\sim$6, 100 & $\times$ & 50.0 \\
        \bottomrule
    \end{tabular}
}
\vspace{0.2cm}
\parbox{\textwidth}{
\footnotesize \textbf{Notes:} 
\textit{Width} denotes the number of requirements that must be satisfied within a single task. \textit{Depth} denotes the number of tool calls/steps per task, where the first value indicates the range of steps taken by the majority of tasks and the second value gives the maximum steps. \textit{Performance} reports GAIA \citep{mialon2023gaia} scores (if available). \textit{Multi-Turn} indicates whether interaction between the user and the agent is multi-turn.
}
\end{table*}

\begin{table*}[h]
\centering
\caption{Comparative Analysis of Environment/Task Diversity across Methods}
\label{tab:diversity}
\resizebox{\textwidth}{!}{
    \begin{tabular}{lccccccccc}
        \toprule
        \multirow{2}{*}{\textbf{Method}} 
        & \multirow{2}{*}{\textbf{Modality}} 
        & \multirow{2}{*}{\textbf{Skill Dimension}} 
        & \multirow{2}{*}{\textbf{\# of Envs}} 
        & \multirow{2}{*}{\textbf{\# of Tasks}} 
        & \multirow{2}{*}{\textbf{\# of Trajs}} 
        & \multirow{2}{*}{\textbf{Source}} 
        & \multicolumn{2}{c}{\textbf{Performance}} \\
        \cmidrule(lr){8-9}
        & & & & & & & \textbf{SWE} & \textbf{ALF} \\
        \midrule
        SWE-Gym-32B \citep{pan2025trainingsoftwareengineeringagents} 
        & Text 
        & Code; 
        & - & 2{,}438 & 491 & Real & 20.6 & - \\
        R2E-Gym-32B \citep{jain2025r2egymproceduralenvironmentshybrid} 
        & Text 
        & Code; 
        & - & 8{,}135 & 3{,}321 & Synth & 34.4 & - \\
        SWE-Smith-32B \citep{yang2025swesmithscalingdatasoftware} 
        & Text 
        & Code; 
        & - & 50{,}137 & 5{,}016 & Both & 40.2 & - \\
        \midrule
        AgentGen-70B \citep{hu2025agentgen} 
        & Text 
        & Planning; 
        & 592 & 11{,}840 & 7{,}246 & Synth & - & 19.4$^{*}$ \\
        \midrule
        AgentGym-70B \citep{xi2024agentgym} 
        & \begin{tabular}[c]{@{}c@{}}Text +\\ Visual\end{tabular} 
        & \begin{tabular}[c]{@{}c@{}}Web; Embodied;\\ Code; Tool Use;\end{tabular} 
        & 14 & 89 & 6{,}130 & Both & - & 88.0 \\
        AgentBank-13B \citep{song2024agentbank} 
        & \begin{tabular}[c]{@{}c@{}}Text +\\ Visual\end{tabular} 
        & \begin{tabular}[c]{@{}c@{}}Web; Embodied;\\ Code; Tool Use;\end{tabular} 
        & 16 & - & 51{,}287 & Both & - & 72.4 \\
        \bottomrule
    \end{tabular}
}
\vspace{0.2cm}
\parbox{\textwidth}{
\footnotesize \textbf{Notes:} \textit{SWE} refers to SWE-bench \citep{jimenez2023swe}. \textit{ALF} refers to ALFWorld \citep{shridhar2020alfworld}. $^{*}$ Performance on out-of-domain tasks.
}
\end{table*}

\section{Evaluation Benchmarks}
\label{sec:eval-bench}
Previous evaluation studies have typically focused on the intelligence of the agents themselves, but there is a lack of direct measurement indicators for aspects such as adaptability to the environment, interactivity, realism, and robustness. Therefore, most environmental assessments are conducted indirectly, usually by observing the performance of intelligent agents to reflect the quality of the environment. For instance, studies such as TaskCraft~\citep{shi2025taskcraft} and AgentScaler~\citep{fang2025generalagentic} train the agents through the trajectories generated by the interaction between the environment and the agents, thereby evaluating the environment. The stronger performance of the agents is regarded as an indirect indication of higher environmental quality. Initially, direct measurements of the environment are mainly limited to symbolic or textual environments. Bytesized32~\citep{wang2023bytesized32} proposes specific-task text games and evaluates them using automated metrics in terms of fidelity, validity, specification adherence, and winnability.  Text2World~\citep{hu2025text2world} benchmarks the generation of symbolic world models, using structural similarity for overall evaluation, and capturing more granular features such as action dynamics through component-level F1 scores. Recent studies have begun to extend the direct assessment to more modalities. VidOSC~\citep{xue2024learning} explores the dynamic characteristics of open-world environments. WorldScore~\citep{duan2025worldscore} proposes a unified framework for evaluating world generation. While WorldPrediction~\citep{chen2025worldprediction} focuses on advanced visual reasoning, emphasizing long-term procedural planning and semantic-time abstraction capabilities. Despite these advancements, comprehensive and universal assessment protocols are still scarce, highlighting the need for more generalized and domain-independent methodologies to rigorously and directly evaluate environmental quality beyond the performance metrics of intelligent agents.

\section{Applications Across Domains}
\label{sec:app}
Driven by interactions with dynamic and diverse environments, recent agentic systems built on state-of-the-art LLM families such as GPT~\citep{openai2025introducing}, Claude~\citep{anthropic2025claude4}, Gemini~\citep{team2024gemini}, and Qwen~\citep{bai2023qwen} have become increasingly powerful and widely applicable.

\paragraph{Tool Use} In tool-use scenarios, environments expose APIs and function calls as structured action spaces for agents, enabling them to reason with external tools through function calling or code generation~\citep{OpenAIFunctionCalling2023,anthropic2024mcp,wu2025webwalkerbenchmarkingllmsweb, mastouri2025makingrestapisagentready,luo2025mcp, wang2025mcp, fan2025mcptoolbench++}. 

\paragraph{Software Engineering} In code generation or software engineering scenarios, environments leverage repositories, test frameworks, and IDE integration to cultivate agents' long-horizon programming capabilities. Systems such as Qwen3-Coder~\citep{qwen3coderblog2024} and Claude 4~\citep{anthropic2025claude4} demonstrate reliability in code editing and debugging. ReTool~\citep{feng2025retoolreinforcementlearningstrategic} novelly integrates code-interpreter execution into the reasoning loop, enabling agents to exhibit code self-correction and adaptive tool selection. SWE-smith~\citep{yang2025swesmithscalingdatasoftware} and R2E-Gym~\citep{jain2025r2egymproceduralenvironmentshybrid} develop scalable, executable code environments based on GitHub repositories, enabling the further evolution of SWE agents.

\paragraph{Web Navigation} In web navigation scenarios, environments are typically built on HTML/DOM structures to support tasks such as browsing, form filling, and transactions~\citep{lai2024autowebglm, AgenticSeek2024, AutoGPT2023}, requiring agents to click buttons and navigate between web pages. Graphical user interface (GUI) environments extend these tasks to desktops and mobile devices, further complicating task complexity~\citep{qin2025uitars,ye2025mobileagentv3,lu2025guiodyssey,liu2024autoglm,sun2022metagui,huang2025guikvefficientguiagents}.

\paragraph{Computer Use} In more recent computer-use scenarios, environments integrate terminals, different operating systems, various applications, and APIs, making tasks much more interactive, open-ended, and realistic~\citep{wang2025opencua, wang2025uitars2, xie2024osworld, OpenAICUA2025,jia2025osworldmcp,xie2025scalingcomputerusegroundinguser}.

\paragraph{Deep Research} In deep research scenarios, environments demand stronger long-context reasoning and more robust retrieval capabilities, as demonstrated by prominent long-form tasks like AstaBench~\citep{bragg2025astabench}, DeepResearchBench~\citep{du2025deepresearchbench}, and ResearchQA~\citep{yifei2025researchqa}. By learning from the interactive experiences collected from these environments, agentic systems such as Tongyi DR~\citep{tongyidr}, Gemini DR~\citep{team2024gemini}, OpenAI DR~\citep{openai2025deepresearch}, and DR Tulu!\citep{shao2025drtulu} are able to conduct sustained, in-depth analysis with mitigated context dilution.

\section{Ethics and Safety Implications}
\label{sec:safety}
Agent systems require robust ethical safeguards, interoperable standards, and alignment with emerging regulatory frameworks to enable responsible integration into the global technology ecosystem \citep{howell2024policy}. Some prior work focuses specifically on safety, such as mitigating inherent biases that produce discriminatory outputs or defending against malicious inputs that circumvent safety-alignment mechanisms \citep{zhao2023art}. However, new challenges are likely to emerge as agents accumulate experience through interactions with their environments and with other agents, and as their capabilities for autonomous task planning, external data access, and tool use continue to grow. For example, prompt injection or tool injection attacks on a single agent may lead to privacy breaches or system-wide contamination, while agents empowered to autonomously operate on external data can, in the absence of strict safeguards, cause irreversible damage by initiating fund transfers or deleting critical files \citep{tang2025risks}. Therefore, when developing scaled environments for future LLM-based agents, safety considerations must be treated as a central and indispensable element of system design.



\end{document}